\documentclass[journal]{IEEEtran}
\ifCLASSINFOpdf
\else
\fi
\usepackage{ulem}
\usepackage{amsmath}
\usepackage{graphicx}
\usepackage{times}
\usepackage{epsfig}
\usepackage{amssymb}
\usepackage{mathrsfs}
\usepackage[table,xcdraw]{xcolor}
\usepackage{color}
\usepackage[colorlinks,linkcolor=blue]{hyperref}
\usepackage{url}
\usepackage{siunitx}
\usepackage{booktabs}
\usepackage{tabularx}
\usepackage{float}
\usepackage{algorithm}
\usepackage{algpseudocode}
\usepackage{amsmath}

\begin{document}

\title{Relation Graph Network for 3D Object Detection in Point Clouds}

\author{Mingtao Feng, Syed Zulqarnain Gilani, Yaonan Wang, Liang Zhang  and {Ajmal Mian}
\thanks{M. Feng and Y. Wang are with the College of Electrical and Information Engineering, Hunan University, Changsha 410082, China (email: mintfeng@hnu.edu.cn; yaonan@hnu.edu.cn).}
\thanks{L. Zhang is with the School of Software, Xidian University, Xi’an 710071, China (email: liangzhang@xidian.edu.cn).}
\thanks{S.Z. Gilani and A. Mian  are with the Department of Computer Science and Software Engineering, The University of Western Australia, Perth, Crawley, WA 6009, Australia (e-mail: {zulqarnain.gilani, ajmal.mian}@uwa.edu.au). S.Z. Gilani is also with the School of Science, Edith Cowan University, Joondalup, 6027, Australia (e-mail: s.gilani@ecu.edu.au)}

\thanks{Manuscript received ***, 2019; revised ***, 2019.}}

\markboth{Submitted to journal , ~~~2019}%
{Shell \MakeLowercase{\textit{et al.}}: Bare Demo of IEEEtran.cls for IEEE Journals}

\maketitle

\begin{abstract}
Convolutional Neural Networks (CNNs) have emerged as a powerful strategy for most object detection tasks on 2D images. However, their power has not been fully realised for detecting 3D objects in point clouds directly without converting them to regular grids. Existing state-of-art 3D object detection methods aim to recognize 3D objects individually without exploiting their relationships during learning or inference. In this paper, we first propose a strategy that associates the predictions of direction vectors and pseudo geometric centers together leading to a win-win solution for 3D bounding box candidates regression. Secondly, we propose point attention pooling to extract uniform appearance features for each 3D object proposal,  benefiting from the learned direction features, semantic features and spatial coordinates of the object points. Finally, the appearance features are used together with the position features to build 3D object-object relationship graphs for all proposals to model their co-existence. We explore the effect of relation graphs on proposals' appearance features enhancement under supervised and unsupervised settings. The proposed relation graph network consists of a 3D object proposal generation module and a 3D relation module, makes it an end-to-end trainable network for detecting 3D object in point clouds. Experiments on challenging benchmarks ( SunRGB-D~\cite{SunRGBD} and ScanNet~\cite{Scannet} datasets ) of 3D point clouds show that our algorithm can perform better than the existing state-of-the-art methods.

\end{abstract}


\section{Introduction}
\IEEEPARstart{W}{ith} the widespread availability of 3D scanning devices, depth sensors and light field cameras \cite{LightField,gilani2018}, 3D point cloud data is being increasingly used in many different application domains such as robotics, autonomous driving, city planning, infrastructure maintenance etc. Accurate detection of 3D objects in point clouds is a central problem for mobile agents to automatically avoid obstacles, plan a route and interact with objects. Converting point clouds to canonical forms such as depth images, multiple views or voxels have been popular approaches to subsequently process the 3D data with Convolutional Neural Networks (CNNs). However, applying CNNs directly on the raw $xyz$ coordinates of the point cloud for 3D object detection has not been widely studied. Progress in 3D object detection lags far behind its 2D counterpart due to the irregular and sparse nature of 3D point clouds. Recently, PointNet~\cite{Pointnet} and PointNet++~\cite{Pointnet++} were proposed to  directly process raw point clouds without converting them to a canonical form. These methods split the input scene into overlapping blocks to avoid the expensive computation and memory cost associated with the huge amounts of data. Unfortunately, this step adversely effects the detection of 3D objects when it is necessary to consider the global scene context.

A straightforward idea is to take inspiration from 2D object detection frameworks to guide the design of 3D methods. For example, Generative Shape Proposal Network~\cite{Gspn} extends the classic 2D-based detector Mask R-CNN~\cite{MaskRCNN} to 3D. It provides an analysis-by-synthesis strategy to generate a large number of 3D object proposals by reconstructing the shapes followed by proposal refinement and instance identification in point clouds. Nonetheless, GSPN~\cite{Gspn} is a dense proposal-based method and relies on two-stage training, which is computationally expensive. More recently, simpler and more efficient 2D object detectors have been proposed ~\cite{Object_as_Point,Centernet,FCOS,Foveabox}. Zhou et al.~\cite{Object_as_Point} represent a 2D object as a single pixel located at the center of its bounding box, and regress the parameters (e.g., dimension, orientation, object size) of each bounding box directly from features of the center pixel. 2D object detection is thus transformed to a standard keypoint detection problem. 

Inspired from the above approach, we extend the 2D object detection scheme and propose an algorithm for accurate geometric center estimation and 3D bounding box regression over 3D point clouds. Specifically, we first introduce a strategy that jointly predicts the pseudo geometric centers and direction vectors  leading to a win-win solution for 3D bounding box candidate regression. A challenge in this cases is that unlike 2D images, where the object's center pixel is surrounded by other pixels, the geometric centers of 3D objects are generally in an empty space\footnote{For example, the center of a sphere is far from the surface of the sphere.} far from points on the object's surface. Moreover, 3D objects in cluttered scenes are usually scanned partially and are noisy. Given the semantic features of points, it is difficult to regress offset values directly that measure the distance from the object surface points to their geometric centers. Inaccurate prediction of pseudo centers will induce error in the downstream 3D bounding box generator. Therefore, to learn more discriminative features from the object surface points and to compute the center position more accurately is the key to regress 3D bounding box candidates. Different from~\cite{VoteNet}, we predict pseudo centers that are close to the geometric center of the object and assign each surface point a direction vector that points to the geometric center. The magnitude and direction of these vectors collaborate to further boosts the accuracy of 3D bounding box candidates. 

Regressing 3D bounding boxes often results in duplicate candidates. A straightforward but naive approach to remove the duplicates is to use 3D Non-Maximum Suppression (NMS) with an Intersection over Union (IoU) threshold. However, an intuitive idea is to exploit the relationship between different objects in the scene. For example, a chair is often close to the desk and a computer is often placed on a desk. This relationship has been exploited in many 2D object recognition algorithms~\cite{RelationNet_CVPR,Actor_relation,Adaptive_connected,Relation_important,Attention_dropout} and is easy to model since related objects are close to each other in images. However, 3D object-object relationships are difficult to model since objects belonging to different categories in cluttered 3D scenes are at arbitrary distances from each other, and their number and sizes are also different.  

We propose an effective 3D relation module that builds a 3D object-object relation graph between 3D bounding box candidates for feature enhancement to achieve accurate object detection. Inspired by~\cite{RelationNet_CVPR}, we introduce a point attention pooling method to extract uniform appearance features for each 3D proposal which are used together with the position features to define nodes of the relation graphs. Since objects in the cluttered scenes are randomly placed and densely connected, (e.g., the seat part of a chair may be under a table) one 3D proposal often contains parts from different objects. Our proposed point attention pooling method exploits the information obtained from 3D proposals by modelling semantic, spacial and direction relationships of the interior points simultaneously. This plays an important role in specifying the intra-object pull forces and the inter-object push forces. 
The above relation graphs are inserted into the main framework and are learned in an unsupervised manner by minimizing the task specific losses, like the 3D bounding box regression loss, cross entropy loss and direction feature loss. 


To sum up, our contributions include: 
\textbf{(1)} A framework for 3D object detection that directly exploits the raw $xyz$ point cloud, is single stage and end-to-end trainable. 
\textbf{(2)} An optimization method which jointly uses the pseudo geometric centers and direction vectors for 3D bounding box candidate estimation. 
\textbf{(3)} A point attention pooling method to extract uniform appearance features for each 3D proposal using semantic features, pseudo geometric centers and direction vectors. 
\textbf{(4)} Constructing a relation graph that exploits the 3D object-object relationships to represent the appearance and position relationship between 3D objects. This enhances the appearance features of each 3D proposal and boosts the performance of 3D bounding box regression. 
We explore the effects of supervised 3D relation graph and multi-graph patterns on 3D relationship reasoning. Experiments are performed on the benchmark SunRGB-D~\cite{SunRGBD} and ScanNet~\cite{Scannet} datasets and achieve state-of-the-art results. We also conduct a series of ablation studies to demonstrate the effectiveness of verious modules of our proposed method.

\begin{figure*}[t!] 
	\center{\includegraphics[width=0.99\textwidth]{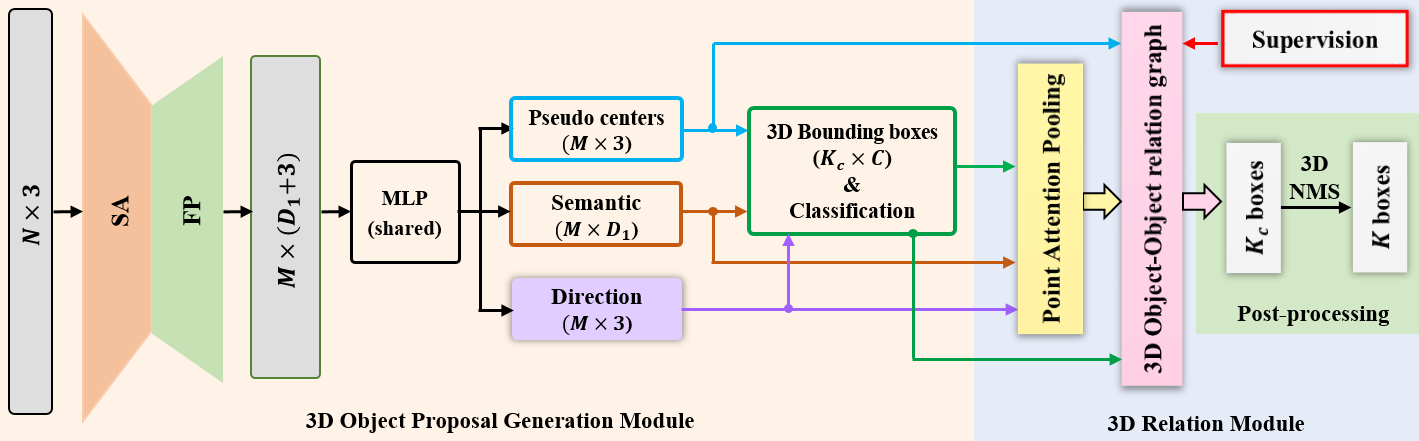}}
	\caption{\label{fig:Framework}Pipeline of our proposed method for 3D object detection in point cloud. The first part is a 3D object proposal generation module to generate 3D bounding box candidates from the raw point cloud using PointNet++\cite{Pointnet++} backbone. The second part is a 3D relation module which contains point attention pooling and 3D object-object relation graph network. The point attention pooling converts the points' features contained in each proposal into uniform vectors, and the 3D object-object relation graph is used to perform relational reasoning on multi graphs or supervised graph which is build on the fixed proposal features. Then the 3D NMS post-processing step is used to pick high quality 3D bounding boxes.
	}
	\vspace{-3mm}
\end{figure*}

\vspace{-3mm}
\section{Related Work}
\textbf{2D Object Detection Methods:} 2D object detection in images is a fundamental problem in computer vision and has been an active area of research for several decades~\cite{Detection_survey}. Numerous methods have been proposed covering different approaches to the problem of generic 2D object detection. Some of these methods can also provide inspirations for 3D object detection in point clouds. Region proposal driven detectors such as RCNN~\cite{RCNN} enumerate object location from region proposals at the first stage, then classify the proposals and refine them at the second stage. Hence, such methods are slow and require a  huge amount of storage. This motivated a series of innovations in this area leading to a number of improved detection methods such as Fast-RCNN~\cite{Fast-RCNN}, SPPNet~\cite{SPPNet}, Faster RCNN~\cite{Faster_rcnn}, RFCN~\cite{RFCN}, Mask RCNN~\cite{MaskRCNN}, Light Head RFCN~\cite{LH_RFCN} etc. 

One-stage detection strategies were introduced which skip the region proposal generation step and directly predict class scores and 2D bounding box offsets from the input images with a single network. Several attempts were made to improve the performance of one-stage detectors, e.g., DetectorNet~\cite{DetectorNet}, OverFeat~\cite{OverFeat}, YOLO~\cite{YOLO} and SSD~\cite{Ssd}. 

Recently, some interesting methods building on robust keypoint estimation networks have been proposed for 2D object detection~\cite{Object_as_Point,Centernet,FCOS,Foveabox,ExtremeNet}. These methods are the inspiration for our proposed 3D bounding box candidates generation method. Specially, Zhou et al.~\cite{Object_as_Point} represent objects by the center pixel of their 2D bounding box. The object centers are obtained by selecting peaks in the heat map generated by feeding the input image to a fully convolutional network. The center keypoint/pixel based 2D object detection methods heavily rely on the heat map and peak pixel estimation whereas it is difficult to generate such heat maps for 3D point clouds. We address this problem by designing a strategy that uses pseudo geometric centers and direction vectors to represent 3D objects, and then regress the 3D bounding box candidates.

\textbf{3D Object Detection Methods:} The most common approach for 3D object detection in a point cloud is to project the point cloud to 2D images for 3D bounding box regression~\cite{Multi_view_3d,Squeezeseg_3d,Single_view_3d}. Point clouds are also sometimes represented by voxel grids for 3D object detection. Zhou et al.~\cite{Voxelnet} divide a full LiDAR point cloud scene into equally spaced 3D voxels and propose a voxel feature encoding layer to learn features for each voxel. Yang et al.~\cite{Pixor} encode each voxel as occupancy and predict oriented 2D bounding boxes in bird’s eye view of LiDAR data. PointPillars~\cite{Pointpillars} organize LiDAR point clouds in vertical columns and then detect 3D objects using a standard 2D convolutional detection framework. 

To avoid voxels and the associated computational cost, Qi et al.~\cite{Frustum} proposed a framework to directly process raw point clouds and then predict 3D bounding boxes based on points within the frustum proposals. However, their algorithm heavily relies on 2D object detection. Moreover, PointRCNN~\cite{PointRcnn} generates 3D bounding box proposals via foreground point segmentation in the first stage, and then learns better local spatial features for box refinement. These methods are designed for 3D object detection in point cloud data obtained from LiDAR sensors. However, LiDAR data is very sparse and there are no cross-connections between different objects that are naturally separate in the 3D space. 

VoteNet~\cite{VoteNet} detects 3D objects in cluttered scenes via a combination of deep point set networks and Hough voting. However, VoteNet is unstable when it comes to voting for the geometric center a partially scanned 3D object. Yang et al.~\cite{Box_instance} extracted a global feature vector through an existing backbone to regress the 3D bounding boxes that ignores small objects and heavily relies on the instance segmentation label. Our proposed 3D bounding box candidates prediction branch is completely different from them as we associate the direction vectors and pseudo geometrical centers for 3D proposal regression. 

\textbf{Networks for Direct Point Cloud Processing.} Learning geometric features directly from point clouds becomes even more essential when color information in unavailable e.g.~in LiDARs. Qi et al.~\cite{Pointnet} proposed PointNet that learns point level features directly from sparse and unordered 3D points. All 3D points are passed through a set of Multi-Layer Perceptrons (MLP) independently and then aggregated to form global features using max-pooling. PointNet achieves promising performance on point cloud classification and segmentation tasks.  The basic PointNet framework has since been extended by many researchers~\cite{Exploring, Pointnet++, DynamicEdge, Local_spectral, Dynamic_condition, PointCNN, RSNet}.  Recently, Duan et al.~\cite{SRN} proposed a structural network architecture for point clouds that takes the contextual information into account. Similarly, Wang et al.~\cite{DynamicEdge} designed a graph convolution kernel that selectively focuses on the most related parts of point clouds and captures the structural features for semantic segmentation. Among these methods, PointNet++~\cite{Pointnet++} is the most commonly used hierarchical framework and is often chosen as the base feature extraction unit for different point cloud related tasks. PointNet++ extracts global features from neighborhood points within a ball query radius, where each local point is processed separately by an MLP. In this work, we use PointNet++~\cite{Pointnet++} as the backbone architecture for point-level feature learning.
 
\section{Proposed Approach}

\subsection{Overview}
Figure~\ref{fig:Framework} shows our framework comprising two parts, one part directly processes the raw points to generate 3D bounding box candidates while the other part  builds the 3D object-object relation graphs to enhance the  appearance features of the proposals for more accurate 3D bounding box regression. Finally, a 3D non-maximum suppression (NMS) is used to remove the duplicate candidates and obtain the final 3D bounding boxes. 

Given a $N\times 3$ point cloud, we first subsample it and learn deep features from it using the PointNet++\cite{Pointnet++}. The output is a subset of $M$ points of size $M\times (D_1+3)$, where $D_1$ is the dimension of learned features. Each subsampled point passes through an MLP with fully connected layers, ReLU and batch normalization and generates a pseudo center, a semantic feature and a direction vector independently. This process enables each sampled point on the object surface to have a direction vector pointing to the geometric center and produces a pseudo center point that is close to the geometric center of the object. To accomplish this task, we propose a direction loss and a cross entropy loss of $K_{sem}$ classes to supervise the network. The pseudo centers, semantic features and direction vectors are then processed to generate $K_c$ 3D bounding box candidates. Based on the direction features and semantic features, we extract uniform appearance features from internal points of each positive proposal using our proposed point attention pooling method. Later, graph convolution networks are used to perform relational reasoning on graph which are built on the appearance and position features. Multi graphs and supervised graph methods are used to enhance the performance of the graph network. Next, the output of graph network is used to enhance the appearance features of the proposals  and regress $K_c$ accurate 3D bounding boxes. Finally, the 3D NMS picks highest quality 3D bounding boxes to output the detected $K$ objects. In the following Sections, we give details of the individual modules of our method.

\begin{figure}[b!] 
	\center{\includegraphics[width=0.45\textwidth]{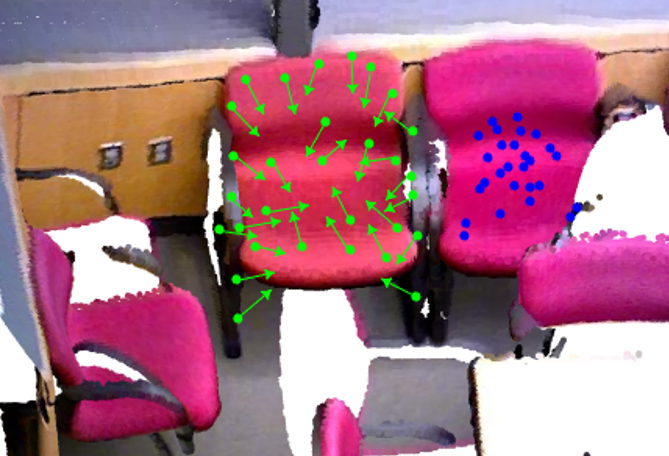}}
	\vspace{-2mm}
	\caption{\label{fig:Direction_fea} Visualization of the direction vectors and pseudo centers. In order to make them easier to distinguish, we select a part of the predicted direction vectors (green arrows) and pseudo centers (blue points) in one point cloud scene and show them in the adjacent two chairs respectively. For better view, we used the color information of the original point cloud, which is not input to our network.}
\end{figure} 

\begin{figure*}[t!] 
	\center{\includegraphics[width=0.99\textwidth]{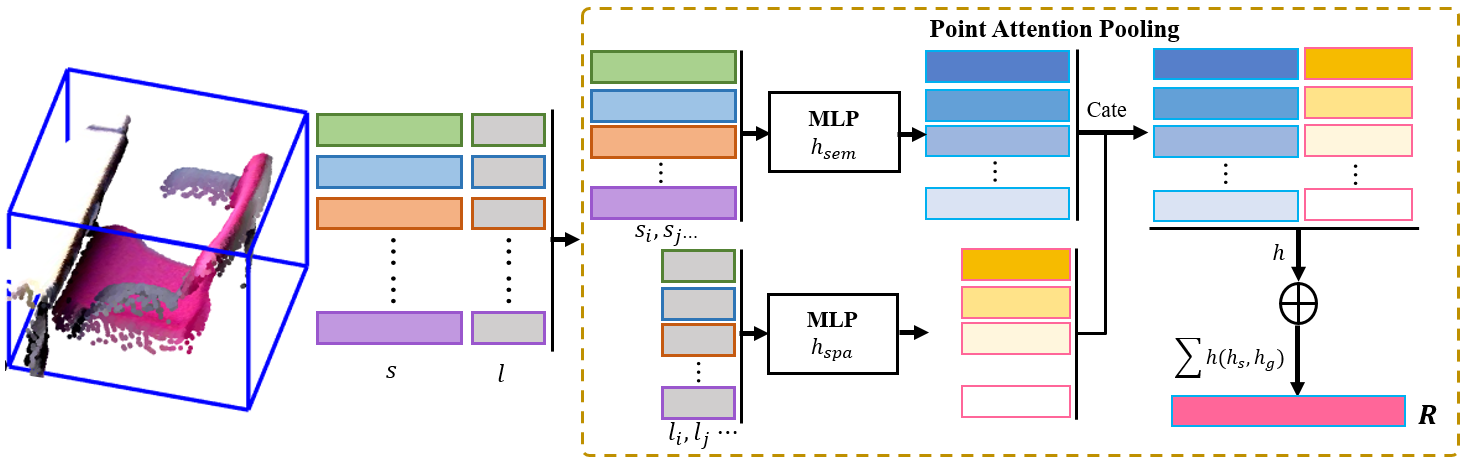}}
	\caption{\label{fig:Proposal_feature} The architecture of proposed point attention pooling branch. Given a 3D bounding box candidate, although the sampled internal points may belong to different objects (desk and chair), we extract the main appearance features of the proposal by separately learning the semantic $h_{sem}$ and spatial interactions $h_{spa}$ because the direction vectors can distinguish the points belonging to different objects. Finally, we extract a fixed appearance feature for each proposal by concatenation, transformation and element wise sum. Colors in this figure have been used to aid a better display.}
	\vspace{-3mm}
\end{figure*}

\subsection{3D Object Proposal Generation}\label{Direction_part}

Inspired by VoteNet~\cite{VoteNet} we first search for the geometric centers of the objects in each semantic class. However, our work significantly differs from VoteNet~\cite{VoteNet} as we propose a new direction loss function to supervise the learning process of the MLP network, obtain the pseudo centers that are close to the geometric center of object and assign each sampled point a direction vector that points to the geometric center of object.

\subsubsection{Direction feature learning}

Given an unordered point cloud $P_0=\left\{p_1,p_2,\ldots,p_N\right\}$ with $p_i\in\mathbb{R}^{d_0}$, where $N$ is the total number of points and $d_0$ is the feature dimension of each point. We use $xyz$ values of the point cloud only i.e. $d_0=3$. 

An entire 3D scene often contains millions of 3D points which are densely sampled on objects that are close to the sensor and sparsely sampled on far objects. Processing these points simultaneously is computationally expensive. Therefore, we subsample the scene to $M$ points ($M << N$) to represent the entire scene. Instead of randomly subsampling the point cloud, we leverage the recently proposed PointNet++~\cite{Pointnet++} for point feature learning due to its efficiency and demonstrated success on tasks ranging from point cloud classification, point cloud semantic segmentation to point cloud generation~\cite{Pointnet,DynamicEdge,PointCNN}. The backbone feature learning network has several Set Abstraction (SA) and Feature Propagation (FP) layers with skip connections, which output a subset of the input points with 3D coordinates $(x,y,z)$ and an enriched $d_1$-dimensional feature vector. The backbone network extracts local point features and selects the most discriminative points within a spherical region. The output $M$ points are denoted by $P_1\in\mathbb{R}^{M\times (d_1+3)}$.

Next, we learn direction vectors pointing to the ground truth geometric center and generate pseudo centers that will be close to the ground truth geometric center for the sampled points $P_1$. Inspired by the concept of center pixels estimation  in 2D object detection~\cite{Object_as_Point}, we regress a 3D bounding box using the predicted pesudo centers and direction vectors jointly. For a set of sampled points $P_1 = \{p_i\}_{i=1}^M$, where $p_i=\left[g_i;f_i\right]$ with $g_i\in \mathbb{R}^3$ and $f_i\in \mathbb{R}^{d_1}$, we train a shared MLP network with fully connected layers, ReLU and batch normalization. The network takes $p_i$ and inputs and outputs the Euclidean space coordinates $g_i' \in \mathbb{R}^3$ and their corresponding feature $f_i'\in \mathbb{R}^{d_1}$ such that the pseudo centers generated from the point $p_i$ are denoted as $p_i'=\left[g_i',f_i'\right]$, $P_1' = \{p_i'\}_{i=1}^M$. The MLP network also outputs a normalized direction feature $v_i\in \mathbb{R}^3$ for each object surface point. We define the vector to be the one pointing towards the ground truth geometric center of each object. The direction feature can describe the inter-object relationship accurately without being affected by other objects. To learn the pseudo center and direction feature, we define the direction loss as follows:
\begin{equation} 
\mathcal{L}_{dir} = \frac{1}{M_{spo}}\left( B_i \sum_{i} \left( \left\|g_i'-g_i^* \right\|- v_i\cdot {v_i}^*  \right) \right)  \label{Dire_loss} 
\end{equation}
where $g_i^*$ is the ground truth geometric center of the 3D bounding box of each object,  $g_i$ is the point on an object surface, $B_i$ indicates whether or not a seed point $p_i$ is on an object surface, $M_{spo}$ is the total number of points on an object surface  and  ${v_i}^*$ is the ground truth normalized direction feature which points towards the geometric center, ${v_i}^*=\frac{g_i-g_i^*}{\left\|g_i-g_i^*\right\|}$ 


Compared to regressing the pseudo centers directly from point semantic features~\cite{VoteNet}, optimizing the direction features and the pseudo centers jointly distributes the estimated pseudo-centers around the geometric center more uniformly. Moreover, the proposed direction loss function generates more discriminate semantic features for points on the object surface in the MLP network  and provides more accurate regional information for subsequent proposal region feature extraction. For illustration purposes, Figure~\ref{fig:Direction_fea} shows the direction vectors (green arrows) and pseudo centers (blue points) generated on surface points of two adjacent chairs in one point cloud scene respectively. We can see that the direction vectors belonging to the same chair  are oriented towards their geometric center, so that different objects can mutually repel each other while different regions belonging to the same object are attracted to each other. Moreover, the pseudo centers cluster at the geometric center of the object providing a basis for the regression of the 3D bounding box candidates together with the direction vectors.

\subsubsection{Proposal Candidates Aggregation}
For a 3D point cloud scene, there are a set of pseudo centers $P_1'\in \mathbb{R}^{3+d_1}$, which create canonical ``meeting points'' for context aggregation from different parts of each object. Similar to VoteNet~\cite{VoteNet}, we sample and cluster these pseudo centers, then aggregate semantic features together with the direction vectors of their corresponding surface points to predict $K_c$ 3D bounding box candidates for all objects and classify them with \textit{objectness scores} and \textit{semantic scores}.  Each proposal is represented by a fixed vector with an objectness score, bounding box parameters $K_b\times C$ ( $C$ represents center, heading and scale parametrized as in~\cite{Frustum}) and semantic classification scores. Refer to the loss function part~\ref{Loss_part} for more details on the parameters.

\subsection{3D Object-Object Relation Graphs}
There is always some relationship between proposals in a 3D scene. We exploit these relationships to enrich the representation of the proposals. However, the modelling of correlation between the proposals faces three primary challenges. First, the points in each region are sparse, varied in number and non-uniformly distributed in space. However, we need to extract fixed dimension appearance features to represent each region, to be used as nodes of the relation graph and play an important role in graph convolution operations. Second, apart from the appearance features, we also need to explore the 3D spatial location interaction between different proposals in order to have sufficient representation capabilities to form the graph nodes. Third, the relationship among proposals is not well defined. Hence, we learn the relationships using multiple graphs or supervised graphs where a center mass loss function is used to guide the relationship learning process.

\subsubsection{Point Attention Pooling}
In a typical 2D object detection pipeline, Region Of Interest (ROI) pooling \cite{Fast-RCNN} or ROI Align~\cite{Faster_rcnn} are used to extract uniform features of each region proposal. However, since the points within a 3D bounding box candidate are usually unordered and sparse, straightforward extension of 2D ROI pooling to point clouds is not possible. We propose a new method named point attention pooling to extract compact features for each 3D bounding box candidate, as shown in Figure~\ref{fig:Proposal_feature}.

\begin{figure*}[t!] 
	\center{\includegraphics[width=0.99\textwidth]{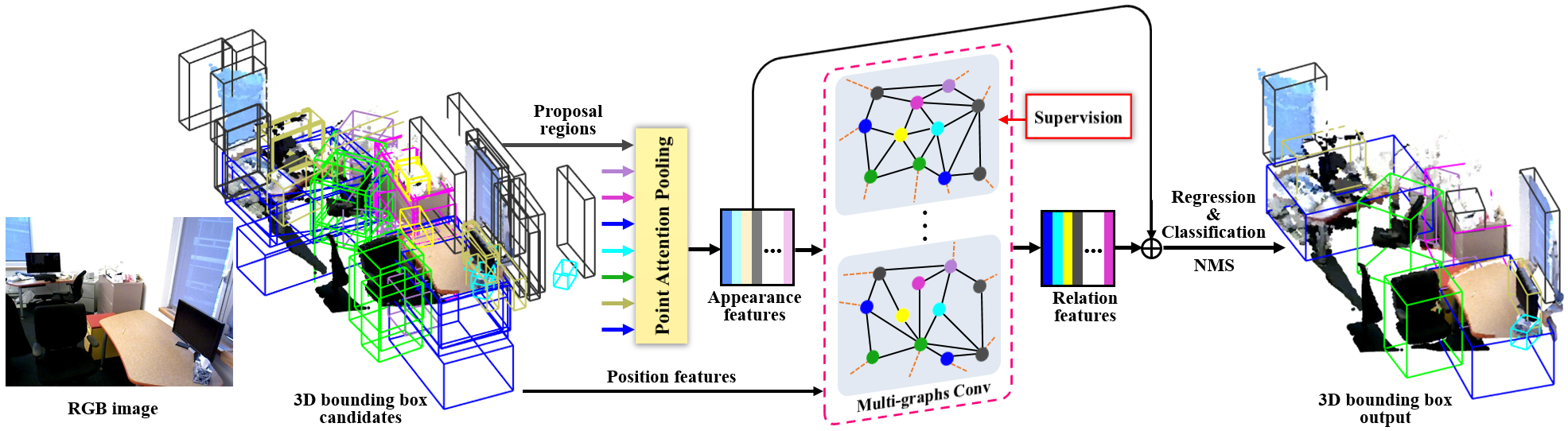}}
	\caption{\label{fig:Graph_object} Overview of our proposed 3D Object-Object Relation Graph. We first extract uniform appearance features for all the 3D bounding box candidates and use them  together with position features to build relation graphs. Next, graph convolutions are conducted to perform relation reasoning. The output of all graphs are fused with the appearance features to regress more accurate 3D bounding boxes. 
		Note that we use only $xyz$ geometric point cloud data for learning and inference. The RGB values are shown only for better visualization.}
	\vspace{-3mm}
\end{figure*}

For each proposal, a naive way would be to apply PointNet++~\cite{Pointnet++} to the interior points without considering their inner interactions and output a uniform feature. However, such an approach does not exploit the semantic information. Instead, our point attention pooling method exploits proposal information by modelling semantic, location and direction relationship of the interior points simultaneously, which plays an important role in indicating intra-object pull forces and inter-object push forces. Our point attention pooling follows two steps. Firstly, we randomly choose $N_{R}$ interior points for each proposal with their semantic features $s\in \mathbb{R}^{d_1}$, spatial coordinates and direction vectors as initial features. When the number of points in a 3D bounding box is less than $N_{R}$, we repeat the interior points until we get the predefined number of points. To make the model robust under geometrical transformations, we move the 3D points of each proposal  to their mean spatial location. The canonical locations are concatenated with the directional vectors to represent spatial features $l\in \mathbb{R}^{6}$ of points within the proposal. In the second step, we explore the semantic and spatial interactions between points $p_i$ and $p_j$. Both semantic features $s$ and spatial features $l$ of interior points play critical roles in interaction learning. For example, repetitive object patterns are exploited by semantic features while the linkage relationship is captured by spatial features. Therefore, we define point attention between the $i$th point and others by jointly learning semantic and spatial interactions:
\begin{equation}
R = \frac{1}{^{N_{R}}}\sum_{j} h\left(h_{sem}\left(s_i,s_j\right),h_{spa}\left(l_i,l_j\right)\right) \label{Point_atten}
\end{equation}
where both $i$ and $j$ are the indices of interior points, $h$, $h_{sem}$ and $h_{spa}$ are functions, and $R\in \mathbb{R}^{1\times d_2}$ is the learned appearance feature for each proposal. The pairwise function $h_{sem}$ and $h_{spa}$ aim to exploit the semantic and spatial relationship between $p_i$ and $p_j$ respectively, and then $h$ fuses the two relationships followed by an element wise sum for all $p_j$. Figure~\ref{fig:Proposal_feature} shows an illustration of the proposed point attention pooling layer, which aims to learn appearance features of the proposal region. Let $d_s$ and $d_l$ be the feature dimensions after the $h_{sem}$ and $h_{spa}$, the number of parameters for point attention pooling are:

\begin{equation}
O(pooling)= K_c(d_2 + d_s + d_l)  \label{Pooling_parameters}
\end{equation}

\subsubsection{Appearance and Position Relationship}
As described above, a series of 3D bounding box candidates are regressed by the pseudo points and direction vectors. Since the 3D scenes are composed of point clouds which are sparse, unordered and usually represent partial objects, the estimated pseudo points will have great uncertainty and can introduce relatively large errors in the regressed 3D bounding box candidates. Once we obtain the uniform appearance features for each 3D bounding box candidate, we explore a method to enhance the features within each proposal region. Inspired by the recent success of relation reasoning and graph neural networks for videos~\cite{Actor_relation}, 2D object detection~\cite{RelationNet_CVPR} and Natural Language Processing (NLP) tasks~\cite{Attention_all}, we use the 3D object-object relation graph structure to explicitly model pairwise relationship information for enhancing the 3D proposal features. To obtain sufficient representational power that captures the underlying relationship between different proposals, both appearance features and position features are considered. Moreover, we note that appearance and position relationships have different effects on the relation graphs. We further investigate this empirically in the ablation study i.e.  section~\ref{Ablation_part}.

Formally, a relation graph is defined as $G=(V,\alpha)$ where $V$ is the set of nodes and $\alpha$ is the set of edges. The nodes in our graph correspond to $K_c$ 3D bounding box candidates and are denoted as $(R_m,U_m)$, where $R_m$ and $U_m$ are appearance features and position features of the $m$-th proposal respectively. We construct the graph $G\in \mathbb{R}^{K_c\times K_c}$ to represent pair wise relationship among the proposals where the relationship value $\alpha_{mn}$ indicates the relative importance of the proposal $n$ to proposal $m$. 

Given an input set of $K_c$ proposals $\{(R_m,U_m)\}_{m=1}^{K_c}$, the relationship feature $F_m$ of  all proposals with respect to the $m$-th proposal is computed as,

\begin{equation}
F_m = \sum_{n} \alpha_{mn} (W_r R_n) \label{Relation_output}
\end{equation}
where $\alpha_{mn}$ is the relationship value between the $m$-th and $n$-th proposals. The output is a weighted sum of appearance features from other proposals, linearly transformed by $W_r$. In our experiment, we adopt the following function to define relation value,

\begin{equation}
\alpha_{mn} = softmax\left( \frac{\alpha_{mn}^P \cdot exp(\alpha_{mn}^A)}{\sum_{K_c} \alpha_{mn}^P \cdot exp(\alpha_{mn}^A)}\right) \label{Relation_value}
\end{equation}
where $\alpha_{mn}^A$ denotes the appearance relationship between two proposals and the position relationship is computed by $\alpha_{mn}^P$. We normalize each relation graph node using softmax function so that the sum of all the relationships for one node is equal to $1$. 
For the appearance relationship, we use dot-product to compute relationship value between two proposals,

\begin{equation}
\alpha_{mn}^A = \frac{(W_a^1R_m)^T(W_a^2R_n)}{\sqrt{d_a}} \label{Dot_product}
\end{equation}
where $d_a$ is transformed feature dimension of the appearance features $R_m$ and $R_n$, and $\sqrt{d_a}$ works as a normalization factor.

For position relationship, the features $U_m = (L_m,S_m)$ represent both the spatial location and geometric structure of each 3D bounding box candidate. The spatial location $L_m$ is represented by the center point of each bounding box while the geometric structure $S_m$ is represented by the parameters of each bounding box. Inspired by~\cite{Actor_relation,Graph_reason}, we investigate two methods to exploit position features for considering the position relationship between proposals: (a) \textbf{3D position mask}. Similar to the image convolution operation where pixels within a local range contribute more to the reference pixel, we assume that proposals from local entities are more important than the proposals from distant entities. Based on the spatial distance between proposals, we define a threshold $\delta$ to filter out distant proposals. Therefore, we set $\alpha_{mn}^P$ to zero for two proposals whose distance is above the threshold. Mathematically, 
\begin{equation}
\alpha_{mn}^P = ReLU \left(W_p \left(\varepsilon(U_m,U_n)|D(L_m,L_n)\le \delta \right)\right) \label{Position_mask}
\end{equation}
where $D(L_m,L_n)$ denotes the Euclidean distance between center points of two proposals and $\delta$ is the distance threshold which is a hyper-parameter. 
The position features are embedded in a high-dimensional representation~\cite{Attention_all} by $\varepsilon$. The feature dimension after embedding is $d_p$. We then transform the embedded features into a scalar by weight vector $W_p$, followed by ReLU activation. (b)\textbf{3D position encoding}. Alternatively, we can use all the proposals to compute their position relationship with the reference proposal. Similar to Equation~(\ref{Position_mask}), the distance threshold is ignored and the rest is retained, as shown below. 

\begin{equation}
\alpha_{mn}^P = ReLU \left(W_p \varepsilon(U_m,U_n)\right) \label{Position_encoding}
\end{equation}

Each relationship function in Equation (\ref{Relation_output}) is parametrized by matrices $W_r$, $W_a^1$, $W_a^2$ and $W_p$. Recall that  $d_2$ is the dimension of the input appearance feature $R$. The number of parameters for one relationship module is

\begin{equation}
O(relation)= K_cd_p + d_a(2d_2 + K_c) + d_2^2 \label{Relation_parameters}
\end{equation}

\subsubsection{Multiple Graphs vs Graph Supervision}
Since the 3D object-object relationship among proposals is not well defined and a single relation graph typically focuses on a specific interaction information between proposals, we extend the single relation graph into multiple graphs to capture complex relationship information. That is, we build the graphs $\mathcal{G}=\left(G^1,G^2,\dots, G^{N_g}\right)$ on the same proposals set, where $N_g$ is the number of graphs. Every graph $G^i$ is computed as in Equation~(\ref{Relation_output}), but with unshared weights. Building multiple relation graphs allows the model to jointly attend to different types of relationships between proposals. Finally, a multi-relation graph module aggregates the total $N_g$ relationship features and augments the input appearance features,

\begin{equation}
R_m = R_m + Sum\{F_m^1,\dots, F_m^{N_g}\}\label{Mutigraphs_output}
\end{equation}

We supervise each graph by giving pseudo ground truth graph weights to learn more accurate relationships. The unsupervised graph weights are learned by minimizing the task specific total loss which contains 3D bounding box regression loss, cross entropy loss, direction feature loss etc. We must construct ground truth labels in matrix form to supervise the learning of $G$ without the need for relationship annotations in the raw point cloud. Our approach is inspired from \cite{FAN}. 
We want our attention weights to focus on relationships between different objects. Hence, for each entry $\alpha_{mn}^T$ of the ground truth relationship label matrix $\alpha^T$, we assign $\alpha_{mn}^T = 1$ only when: (1) 3D object $m$ and 3D object $n$ overlap with the ground truth 3D bounding boxes of two different objects with IOU $\geq0.15$ and (2) their category labels are different.

\begin{equation}
L_{sup} = -\left(1-M\right)^2log(M) \label{Supervision_relation}
\end{equation}
where $M=\sum softmax(G) \odot G^T$ is the center of mass. When minimizing this loss, we would like $G$ to have high relation weights at those entries where $\alpha_{mn}^T = 1$, and low relation weights elsewhere.  

As shown in Figure~\ref{fig:Graph_object}, the appearance features are first extracted from 3D bounding box candidates and then used with the position features to build relation graphs. Graph convolution is then used to perform relation reasoning. The outputs of all graphs are then fused with the appearance features to regress more accurate 3D bounding boxes. We use the multiple graphs and graph supervision methods to explore which one is more beneficial to the establishment of relationships between different proposals. Their performance is discussed in the ablation study~\ref{Ablation_part}. 

\subsection{Loss Function}\label{Loss_part}
Our complete network can be trained in an end-to-end manner with a multi-task loss including a directional loss, an objectness loss, a 3D bounding box estimation loss and a semantic classification loss. We weigh the losses such that they are in similar scales with $\lambda_1=0.4$, $\lambda_2=1$,  $\lambda_3=0.2$, and $\lambda_4=0.1$ when we use supervised graph model or $\lambda_4=0$.

\begin{equation}
L_{total} = L_{dir} + \lambda_1L_{obj} + \lambda_2L_{box} + \lambda_3L_{sem} + \lambda_4L_{sup} \label{Loss_total}
\end{equation}
The direction regression loss $L_{dir}$ is defined in Equation~\ref{Dire_loss} and discussed in detail in Section~\ref{Direction_part}. 
Note that the SUN RGB-D~\cite{SunRGBD} dataset does not provide instance segmentation annotations. Therefore, we compute the ground truth object centers as the centers of the 3D bounding boxes and consider any point inside a ground truth bounding box as an object point. Similar to~\cite{VoteNet}, we keep a set of up to three ground truth votes, and consider the minimum distance between the predicted vote and any ground truth vote in the set during vote regression on this point. For ScanNet~\cite{Scannet}, we consider any point sampled from instance mesh vertices as an object point and compute the ground truth object center as 3D bounding box center.

The objectness loss $L_{obj}$ is a cross-entropy loss for two classes (positive and negative proposals) while the semantic classification loss is the cross-entropy loss for $C$ classes. We follow~\cite{VoteNet,Frustum} in defining the box loss, which comprises the center regression, heading estimation and size estimation sub-losses. Specifically, $L_{box} = L_{c-reg} + 0.1L_{h-cls} + L_{h-reg} + 0.1L_{s-cls} + L_{s-reg}$, where $L_{c-reg}$ is the loss for the box center regression, $L_{h-cls}$ and $L_{h-reg}$ are losses for heading angle estimation while $L_{s-cls}$ and $L_{s-reg}$ are losses for bounding box size regression. The dimension of the output of the last layer is $2 + 6 + 2NH + 4NS + NC$ channels, where the first $2$ channels are for objectness classification, the $6$ channels are for pseudo center and directional vector regression, $NH$ is the number of heading bins, $NS$ is the number of size templates and $NC$ is the number of semantic classes. We use the robust $L_1$ smooth loss in all regressions for the box loss. Both the box and semantic losses are only computed on positive vote clusters and normalized by the number of positive clusters.

\subsection{Comparison to 2D Visual Relationships}
The classic 2D object detection approaches, e.g., RCNN~\cite{RCNN}, Fast-RCNN~\cite{Fast-RCNN} and Faster RCNN~\cite{Faster_rcnn}, only use features within the proposals to refine the bounding boxes. The surrounding and long term information is not considered which is also important for 2D object detection. Santoro et al.~\cite{Relation_reason} introduced a relation networks augmented method on visual question answering between different objects and achieved performance superior human annotators. Hu et  al.~\cite{RelationNet_CVPR} proposed an object relation module to learn the relationships between different proposals which captures the 2D appearance and location relations simultaneously, and evaluated the effectiveness of inserting the modelled relations in the RCNN based detection framework. Wu et al.~\cite{Actor_relation} used actor relation graphs to learning the relation information between multi person for recognizing group activity, which achieved significant gain on group activity recognition accuracy. Moreover, many works also showed that modelling relation information are useful for action recognition~\cite{Non_local,Acentric_relatioin,Interact_video}. These image based relation models mostly rely on the extraction of regions of interest (RoI) features where regular pooling methods can be used, and the definition of location relation generally use the center point of each 2D bounding box since the 2D objects in an image are interlaced and occluded. Although these methods are instructive for our 3D relationships learning for object detection in point cloud, they can not be used directly. Therefore, we design a completely new approach to establish interaction model between 3D objects.

In 3D point cloud scenes, objects of various sizes are randomly placed in space, often occluding each other and with dense object-to-object connections. Qi et al.~\cite{VoteNet} proposed a VoteNet which detects 3D objects from the raw point clouds without splitting the scene into overlapped cubes. VoteNet regresses the 3D bounding boxes for all 3D objects using voting, sampling and clustering. However, calculating the pseudo centers directly from the sparse and unordered points on object surface is an unstable approach and will affect the regression of 3D bounding boxes from pseudo centers. Yi et al.~\cite{Gspn} introduced a generative shape proposal network (GSPN) for 3D instance segmentation which takes an analysis-by-synthesis strategy to generate 3D proposals for all instances where the shape proposal generation is just an intermediate process. Similar to GSPN~\cite{Gspn}, Yang et al.~\cite{Box_instance} segment the instances in 3D point cloud scene by regressing 3D bounding boxes for all instances. The proposed 3D-BoNet extracts a global feature vector through an existing backbone to regress the 3D bounding boxes, which ignores small objects. Moreover, the GSPN~\cite{Gspn} and 3D-BoNet~\cite{Box_instance} heavily rely on the point level mask labels. Note that all these methods do not consider the relationships between the surrounding objects and semantic information in the global 3D space. Unlike the above frameworks, we propose the relation graph network to detect 3D objects in point cloud scenes. We regress the 3D bounding box candidates through the predicted pseudo centers and direction vectors jointly, where these two features can take advantage of each other to further boost the accuracy of 3D bounding box candidates. We also build the 3D object-object relation graph module using appearance features and position features to learn the interactions between different 3D proposals for 3D bounding box refinement. Furthermore, we explore multi graphs and supervised graph strategies to drive relation modules that learn stronger relationships.

\section{Experiments and Discussion}
We first introduce two widely-used 3D object detection benchmarks and the implementation details of our method and then present a series of ablation studies to analyse the efficacy of the proposed units in our model. We also compare the performance of our method with the state of art. Finally, we show visualizations of our learned 3D object-object relation graph and present the 3D object detection results.

\subsection{Dataset}
All experiments are performed on the publicly available SunRGB-D~\cite{SunRGBD} and ScanNet~\cite{Scannet}  datasets. The SunRGB-D~\cite{SunRGBD} dataset contains 10,335 RGB-D images with dense annotations in both 2D and 3D for all object classes. We split it into a train set of 7,000 scenes and a validation set of 3,335 scenes. For our purpose, we reconstruct point cloud scenes from the depth images using camera calibration parameters, where each object is annotated by a 3D bounding box presented by center coordinates, orientations and dimensions. 

The ScanNet~\cite{Scannet} dataset contains 1,513 scans of about 707 unique real-world environments. A ground truth instance level semantic label is assigned to each reconstructed 3D surface mesh. We split the data into a train set of 1200 scans and a validation set of 3,335 scans. We sample points from vertices of the 3D surface mesh and compute 3D bounding box of each instance following the method proposed by~\cite{3DSIS}.

Following~\cite{STDdetector,VoteNet}, we augment the training data by randomly flipping each point cloud scene   along the $X$-axis and $Y$-axis in camera coordinates, randomly rotating around the $Z$-axis by an angle selected uniformly between $[-30^\circ, 30^\circ]$ and globally scaling between $[0.9, 1.1]$. We follow the standard protocols for performance evaluation.

The SunRGB-D~\cite{SunRGBD} and ScanNet~\cite{Scannet} datasets are mainly indoor scenes where the objects are densely interlaced and randomly placed. A 3D bounding box may contain one object and partial areas from other objects in some cases. Limited by the RGB-D sensors, the SunRGB-D~\cite{SunRGBD} has partial scans and the reconstructed point cloud scenes are noisy. These conditions make it challenging to detect 3D objects directly from point clouds. Compared to the SunRGB-D~\cite{SunRGBD}, ScanNet~\cite{Scannet} contains more complete objects. 

\subsection{Implementation Details} 
Our 3D object detection architecture contains a 3D proposal generation module and a 3D relationship module followed by 3D NMS post-processing. 
In the first stage, we randomly sample $N=20K$ points from each reconstructed point cloud scene of SunRGB-D~\cite{SunRGBD} and $N=35K$ points from the 3D scans of ScanNet~\cite{Scannet}.The 3D object proposal generation module is based on PointNet++~\cite{Pointnet++} with four Set Abstraction (SA) layers for learning local features and two Feature Propagation (FP) layers for upsampling. The receptive radius of the SA layers are $0.2$, $0.4$, $0.8$ and $1.2$ in meters, the number of the subsampled points are $2048$, $1024$, $512$ and $256$ respectively. The output size of PointNet++ is $1024\times 259$, where $M=1024$ is the number of sub-sampled points and $259$ is the feature dimension (the last $3$ channels are for 3D coordinates). The data is then fed to an MLP and the size of outputs are $1024\times259$, $1024\times259$ and $1024\times262$ respectively, the last $6$ channels are for 3D coordinates of pseudo centers and direction vector of sampled $M$ points. After sampling and clustering, a total of $K_c=256$ 3D bounding boxes are generated. 

In the second stage, we extract a fixed dimensional appearance feature $R_m=1\times256$ for constructing the 3D object-object relation graph. We arbitrarily set the distance threshold $\delta$ to $\frac{1}{4}$ of the $\sqrt{(d_x^2+d_y^2+d_z^2)}$, where $d_x$, $d_y$ and $d_z$ are the $xyz$-axis limits of each 3D scene. We set $N_R=128$, $d2=256$, $d_s=64$, $d_l=32$, $d_p=256$ and $d_a=256$. In SUN RGB-D: $NH=12,NS = NC = 10$, in ScanNet: $NH=12, NS = NC = 18$.

We train the entire network end to end from scratch with an Adam optimizer, batch size 6 and initial learning rate of $0.001$. The learning rate is decreased by a factor of $10$ after $100$ and $160$ epochs. We train the network on two NVIDIA GTX TitanX GPUs in the PyTorch framework. Our network is able to take point clouds of entire scenes and generate proposals in one forward pass.

\begin{table}
	\centering
	\caption{\label{Direction_effect} Exploring the effect of direction features. Evaluation metric is mean Average Precision (mAP) with 3D bounding box IoU threshold 0.25 as proposed by~\cite{SunRGBD}.}
	\setlength{\tabcolsep}{3mm}{
	\begin{tabular}{l|c}
		\toprule
		Method                     & mAP@0.25 \\
		\midrule 
		Base model                  & 57.2     \\ 
		\midrule 
		Base model + Direction features (ours) & \textbf{57.9}     \\ 
		\bottomrule
	\end{tabular}}
\end{table}

\begin{table}
	\centering
	\caption{\label{Pooling_effect}Analysing the effect of the proposed point pooling attention method. We compare the point attention pooling with SA layer~\cite{Pointnet++}, feature average and feature maximum. We also evaluate the effect of direction features on point attention pooling method.}
	\setlength{\tabcolsep}{4.5mm}{
		\begin{tabular}{l|c}
			\toprule
			Methods                & mAP@0.25 \\ 
			\midrule
			SA layer~\cite{Pointnet++}               & 58.6     \\ 
			Feature average           & 58.0     \\ 
			Feature maximum            & 57.6     \\
			\midrule
			Point attention pooling w/o direction features & 58.9 \\
			Point attention pooling (ours) & \textbf{59.2}     \\ 
			\bottomrule
	\end{tabular}}
\end{table}

\subsection{Ablation Studies}
\label{Ablation_part}
We conduct four groups of ablation experiments on the SunRGB-D~\cite{SunRGBD} dataset. We select this dataset as it contains partial and noisy scans thereby making our task more challenging.

\subsubsection{Effect of Direction Features}
To simplify the experiments and get more intuitive result, we build a concise base framework that only contains the 3D proposal generation module and 3D NMS post-processing. Hence, the total loss function of the base model are defined as $L_{total} = L_{dir}' + \lambda_1L_{obj} + \lambda_2L_{box} + \lambda_3L_{sem}$, where $\lambda_1=0.5$, $\lambda_2=1$ and $\lambda_3=0.1$, the direction loss function $L_{dir}' = \frac{1}{M_{spo}} B_i \sum_{i} \left\|g_i'-g_i^* \right\|$, which only contains the distance norm between the sampled points on object surface and their corresponding object geometric center. In contrast to $L_{dir}'$, we use the proposed direction loss function $L_{dir}$, as defined in Equation (\ref{Dire_loss}), which also generates a direction vector for each object surface point while regressing a pseudo center for them. The settings of PointNet++ remain the same as the original network. 

As shown in Table~\ref{Direction_effect}, using the proposed loss function results in improved performance which means that the direction features improve pseudo center estimation and 3D bounding boxes regression. 


\subsubsection{Point Attention Pooling}
To extract appearance features from each 3D bounding box candidate for relation graph construction, we propose a point attention pooling method to make use of the interactions of the interior points. In this section, we explore the efficacy of point attention pooling as compared to set abstraction (SA) layer~\cite{Pointnet++}, feature average or feature max. Feature average and feature maximum methods are connected to the MLP and their outputs are one dimensional features of the same size as our proposed method. We use these four methods to extract proposal features and then use them as appearance features to build relation graph. The other settings of our proposed framework remain the same as the original network.

\begin{figure} 
	\center{\includegraphics[width=0.45\textwidth]{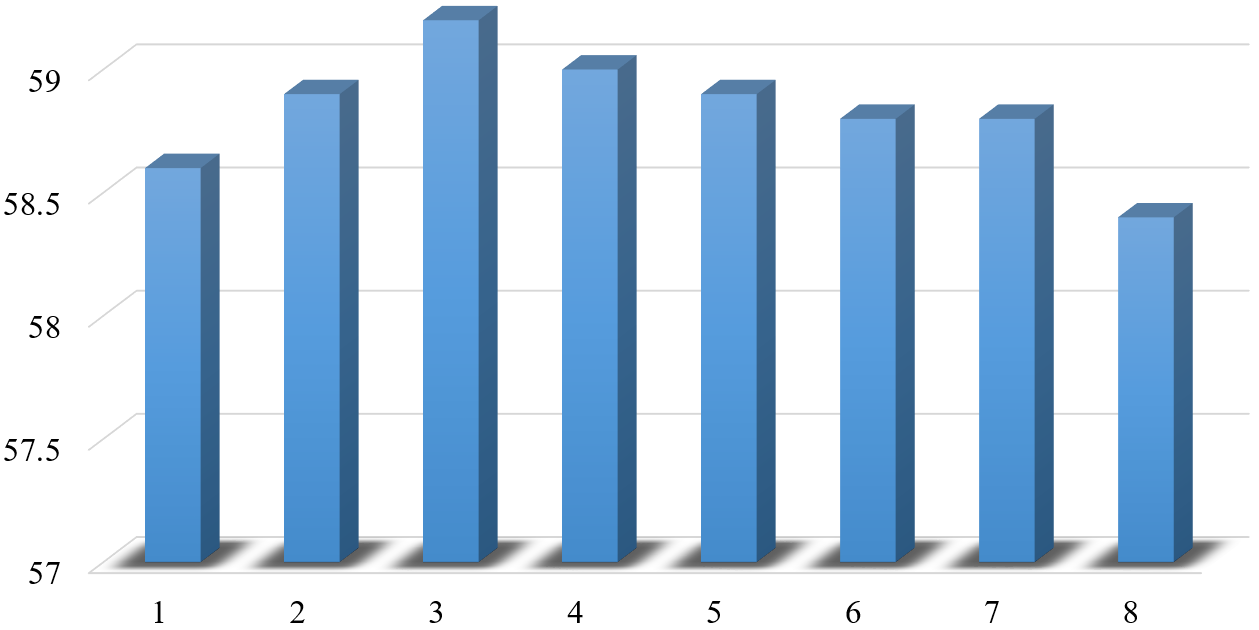}}
	\vspace{-2mm}
	\caption{\label{Graph_number} Comparing the affect of varying number of graphs on 3D object detection performance keeping the other settings of our framework unchanged.}
\end{figure}   

\begin{table}
	\centering
	\caption{\label{Graph_effect} Exploring the effect of 3D object-object relation graph on 3D object detection. We mainly consider two factors, i.e.,  the number of relation graphs and supervised relation graph.}
	\setlength{\tabcolsep}{6mm}{
		\begin{tabular}{l|c}
			\toprule
			Methods                      & mAP@0.25       \\ 
			\midrule
			w/o graph                    & 57.9           \\
			1 graph w/o supervision      & 58.3           \\  
			1 graph w/ supervision       & 58.5           \\
			\midrule
			3 graphs w/o supervision     & \textbf{59.2}  \\ 
			\bottomrule
	\end{tabular}}
\end{table}

\begin{table}
	\centering
	\caption{\label{Feature_effect} Comparison on the effect of appearance and position features. }
	\setlength{\tabcolsep}{3mm}{
		\begin{tabular}{l|c}
			\toprule
			Method                 & mAP@0.25 \\ 
			\midrule
			w/o appearance features & 57.5     \\ 
			w/o position features  & 58.4     \\ 
			w/ appearance features  (w/ 3D postion mask)        & 58.9     \\ 
			w/ appearance features (w/ 3D position encoding)   & \textbf{59.2}     \\ 
			\bottomrule
	\end{tabular}}
\end{table}

From Table~\ref{Pooling_effect} we can see that the point attention pooling method achieves the best performance. We observe that the feature average and feature maximum methods do not perform as good because they extract proposal region features by simply computing the mean and extreme values respectively. 
Since objects in the scenes of SunRGB-D~\cite{SunRGBD} dataset are randomly placed and densely connected, e.g., the seat part of a chair may be under a table, a 3D bounding box candidate often contains parts from different objects. Therefore, it is necessary to fully gather the points of the same object  and learn the semantic and geometric information associated with  them when extracting proposal region features. In the last two rows of Table~\ref{Pooling_effect}, we can see that direction features on the point attention pooling does provide some improvement. Our interpretation is that the direction vectors make points of the same object attract each other, and points of different objects repulse each other.

\begin{table*}
	\caption{\label{SunRGBD_Acc} Quantitative comparison of 3D object detection in point cloud on SunRGB-D~\cite{SunRGBD} dataset. Evaluation metric is mean average precision (mAP) with 3D bounding box IoU threshold 0.25 as proposed by~\cite{SunRGBD}. }
	\resizebox{\textwidth}{16mm}{
		\begin{tabular}{l|c|c|c c c c c c c c c c}
			\toprule
			Methods & Input    &mAP@0.25     & bathtub  & bed  & bookshelf & chair& desk & dresser & nightstand & sofa & table & toilet   \\ 
			\midrule
			DSS~\cite{DSS}        & RGB+Points  & 42.1 & 44.2 & 78.8      & 11.9 & 61.2  & 20.5 & 6.4  & 15.4 & 53.5 & 50.3  & 78.9    \\ 
			COG~\cite{COG}        & RGB+Points  & 47.6 & 58.3 & 63.7      & 31.8 & 62.2  & \textbf{45.2} & 15.5 & 27.4 & 51.0 & \textbf{51.3}  & 70.1    \\ 
			2D-driven~\cite{2d_driven}  & RGB+Points  & 45.1 & 43.5 & 64.5      & 31.4 & 48.3  & 27.9 & 25.9 & 41.9 & 50.4 & 37.0  & 80.4    \\ 
			F-PointNet~\cite{Frustum} & RGB+Points  & 54.0 & 43.3 & 81.1      & \textbf{33.3} & 64.2  & 24.7 & \textbf{32.0} & 58.1 & 61.1 & 51.1  & 90.9    \\ 
			VoteNet~\cite{VoteNet}    & {Points only} & 57.7 & 74.4 & 83.0      & 28.8 & 75.3  & 22.0 & 29.8 & 62.2 & 64.0 & 47.3  & 90.1    \\ 
			\midrule
			\textbf{Ours}        & \textbf{Points only} & \textbf{59.2}     & \textbf{75.2} & \textbf{83.7} & 31.2 & \textbf{76} & 28.1 & 	29.4 &	\textbf{62.8} &	\textbf{66.5} &  48 & \textbf{91}
			              \\ 
			\bottomrule
	\end{tabular}}
\end{table*}

\begin{table*}
	\centering
	\caption{\label{ScanNet_Acc_detail} 3D object detection scores per category on the ScanNet~\cite{Scannet} dataset in terms of the mAP@0.25 and mAP@0.5 IoU respectively. }
	\setlength{\tabcolsep}{0.82mm}{
		\begin{tabular}{l|c|c|c|c|c|c|c|c|c|c|c|c|c|c|c|c|c|c|c}
			\toprule
			Method                           & mAP@0.25              & cab & bed & chair & sofa & tabl & door & wind & bkshf & pic & cntr & desk & curt & fridg & showr & toil & sink & bath & ofurn \\ 
			\midrule
			3D-SIS 5views\cite{3DSIS}  & 40.23   & 19.76 & 69.71 & 66.15 & 71.81 & 36.06 & \textbf{30.64} & 10.88 & 27.34 & 0.0 & 10.0 & 46.93 & 14.06 & 53.76 & 35.96 & 87.60  & \textbf{42.98} & \textbf{84.3} & 16.2         \\ 
			3D-SIS points\cite{3DSIS}                       & 25.36 & 12.75 & 63.14 & 65.98 & 46.33 & 26.91 & 7.95 & 2.79 & 2.3 & 0.0 & 6.92 & 33.34 & 2.47 & 10.42 & 12.17 & 74.51 & 22.86 & 58.66 & 7.05 \\ 
			VoteNet\cite{VoteNet}  & 46.8 & 29.96 & 82.87 & 78.66 & 76.32 & 53.1 & 29.64 & 23.57 & 33.59 & 2.37 & 36.92 & 57.76 & 33.71 & 33.68 & 45.14 & 86.98 & 37.96 & 79.77 & 19.58 \\ 
			\midrule
			Ours    &  \textbf{48.5} & \textbf{31.08} & \textbf{83.1} & \textbf{85.86} & \textbf{77.5} & \textbf{56.27} & 30.55 & \textbf{25.1} & \textbf{34.84} & \textbf{4.09} & \textbf{38.5} & \textbf{59.11} & \textbf{35.32} & \textbf{33.7} & \textbf{46.29} & \textbf{88.6} & 40.27 & 82 & \textbf{20.9}
			       \\
			\midrule
			\midrule
			Method                       &mAP@0.5               & cab & bed & chair & sofa & tabl & door & wind & bkshf & pic & cntr & desk & curt & fridg & showr & toil & sink & bath & ofurn \\ 
			\midrule
			3D-SIS 5views\cite{3DSIS}  & 22.53 & 5.73 & 50.28 & \textbf{52.59} & \textbf{55.43} & 21.96 & \textbf{10.88} & 0.0 & 13.18 & 0.0 & 0.0 & 23.62 & 2.61 & \textbf{24.54} & 0.82 & \textbf{71.79} & 8.94 & 56.40 & \textbf{6.87}   \\ 
			3D-SIS points\cite{3DSIS} & 14.6 & 5.06 & 42.19 & 50.11 & 31.75 & 15.12 & 1.38 & 0.0  & 1.44 & 0.0 & 0.0 & 13.66 & 0.0 & 2.63 & 3.0 & 56.75 & 8.68 & 28.52 & 2.55  \\ 
			VoteNet\cite{VoteNet}     & 24.7 & 6.42 & 71.26 & 45.14 & 50.19 & 31.88 & 6.15 & 4.04 & 22.32 & 0.08 & 5.14 & 21.32 & 9.36 & 14.27 & 7.75 & 65.56 & 16.5 & 62.86 & 3.53  \\ 
			\midrule
			\textbf{Ours}     & \textbf{26.0} &\textbf{ 7.39} & \textbf{72.42} & 49.84 & 51.8 & \textbf{33.5} & 6.97 & \textbf{5.73} & \textbf{23.11} & \textbf{0.15} & \textbf{5.94} & \textbf{24.91} & \textbf{10.2} & 14.95 & \textbf{8.09} & 67.79 & \textbf{17.16} & \textbf{63.74} & 4.17
			      \\ 
			\bottomrule
	\end{tabular}}
\end{table*}

\begin{table}
	\centering
	\caption{\label{ScanNet_Acc} Quantitative comparison of 3D object detection in point cloud on ScanNet~\cite{Scannet} dataset. Evaluation metric is mean average precision (mAP) with 3D bounding box IoU threshold 0.25 and 0.5.}
	\setlength{\tabcolsep}{2.5mm}{
		\begin{tabular}{c|c|c|c}
			\toprule
			Methods & Input          & mAP@0.25 & mAP@0.5 \\ 
			\midrule
			DSS~\cite{DSS}        & RGB+Points     & 15.2     & 6.8     \\ 
			MRCNN~\cite{3DSIS,MaskRCNN}      & RGB+Points     & 17.3     & 10.5    \\ 
			F-PointNet~\cite{Frustum} & RGB+Points     & 19.8     & 10.8    \\ 
			GSPN~\cite{Gspn}       & RGB+Points     & 30.6   & 17.7    \\ 
			3D-SIS~\cite{3DSIS}     & 5 views+Points & 40.2     & 22.5    \\ 
			3D-SIS~\cite{3DSIS}     & \textbf{Points only}    & 25.4     & 14.6    \\ 
			VoteNet~\cite{VoteNet}    & \textbf{Points only}    & 46.8     & 24.7    \\ 
			\midrule
			Ours       & \textbf{Points only}    & \textbf{48.5}     & \textbf{26.0}    \\ 
			\bottomrule
	\end{tabular}}
\end{table}

\subsubsection{3D Object-object Relation Graph}

We now perform ablation studies on the following three key parameters: 

(a) Number of relation graphs: As shown in Figure~\ref{Graph_number}, using more relation graphs (while keeping everything else constant) steadily improves the accuracy of 3D object detection up to three graphs after which there is a gradual drop in accuracy. Therefore, we use three graphs in the remaining experiments unless otherwise mentioned. One might intuitively think that the 3D object detection accuracy will continue to improve with increasing number of graphs but practically, this is not the case. The relationships between different objects are learned by building graphs using fixed size proposal region features, which enhances the features of each region to return more accurate 3D bounding boxes. The relationships between different objects in most scenes of SunRGB-D~\cite{SunRGBD} dataset are not particularly complicated, and although augmentation is performed, the training data is limited to only 7000 scenes. When the number of graphs increases, the number of model parameters also increases leading to over-fitting. 

(b) Supervised relation graph: To examine the supervised graph strategy for 3D object detection, we choose a baseline model with $1$ graph only  and an improved model with $1$ graph supervised by the center of mass loss. All other settings of the framework were kept fixed during this experiment. The results are given in Table~\ref{Graph_effect}, and indicate that the supervised graph method is indeed better than that unsupervised case. 
In addition, we observe that the $3$ graphs model without supervision is better than the $1$ graph model with supervision. 
The center mass loss function allows the network to directly learn the relationships between different objects that are predefined by us. In the absence of supervision, the relationship coefficients between different objects are indirectly learned in conjunction with the task specific loss function. The detection results depend more on the size of the training data and the design of the network structure. Detection result of the supervised $1$ graph model is worse than the $3$ graphs model because we only provide one fixed prior situation which can only reflect one relationship case between different objects. Although the $3$ graphs model is supervised by the task specific loss function, it has enough parameter space to explore the relationships between different objects and hence works the best. Therefore, in the remaining experiments, we use a $3$ graphs model without supervision unless stated otherwise.

(c) Usage of appearance and position features: We firstly study the effect of appearance features on modelling the 3D object-object relation graphs abstracted by our proposed point attention pooling method. We build a single framework without using appearance features to build the relation graphs. The results are listed in the first and third rows of Table~\ref{Feature_effect}, it is obvious that explicitly modelling the relation graphs between different objects using appearance features improves performance. Next, we study the effect of position features on modelling the relation graphs, which are defined by the distance mask and distance encoding. From the last three rows of Table~\ref{Feature_effect}, we observe that the position features yield improvement for 3D object detection accuracy, and the distance mask performs better than the distance encoding. The appearance features extracted from each proposal region play an important role in the process of building relation graphs, which is used to represent 3D bounding box candidates discriminatively. The use of appearance features determines whether the relation graphs can learn the relationships between different objects and enhance the features of each proposal region. Due to the complexity of spatial distribution within the scenes, the introduction of position features also better establishes the spatial relationships between different objects. Thus, in all our experiments, we apply the appearance and position features together to build relation graphs unless stated otherwise.

\subsection{Complexity}

We use the following values for the parameters in Equations (\ref{Pooling_parameters}) and (\ref{Relation_parameters}), $K_c=256$, $d_2=256$, $d_s=64$, $d_l=32$, $d_p=256$, $d_a=256$. When $N_g=3$, our proposed point attention pooling method has about 0.09 million parameters and the relation graphs module has about 0.9 million parameters. Table~\ref{Model_size} shows that the complexity of our proposed network is much lower than F-PointNet~\cite{Frustum} and 3D-SIS~\cite{3DSIS}, but a little more than VoteNet~\cite{VoteNet}. The base model indicates the most simple pipeline that does not include the point attention pooling and 3D object-object relation graph, and is similar to VoteNet~\cite{VoteNet}. We can see that the increase  in model complexity brought by our complete framework is relatively small compared to the entire detection architecture.

\begin{table}
	\centering
	\caption{\label{Model_size} Comparison of the model size for different methods.}
	\setlength{\tabcolsep}{6mm}{
		\begin{tabular}{l|c}
			\toprule
			Methods    & Model Size  \\
			\midrule 
			F-PointNet~\cite{Frustum} & 47.0M            \\ 
			3D-SIS~\cite{3DSIS}       & 19.7M         \\ 
			VoteNet~\cite{VoteNet}    & 11.2M         \\ 
			\midrule
			Our base model & 11.25M          \\
			Ours       & 12.1M         \\ 
			\bottomrule
	\end{tabular}}
\end{table}

\begin{figure}[t!] 
	\center{\includegraphics[width=0.5\textwidth]{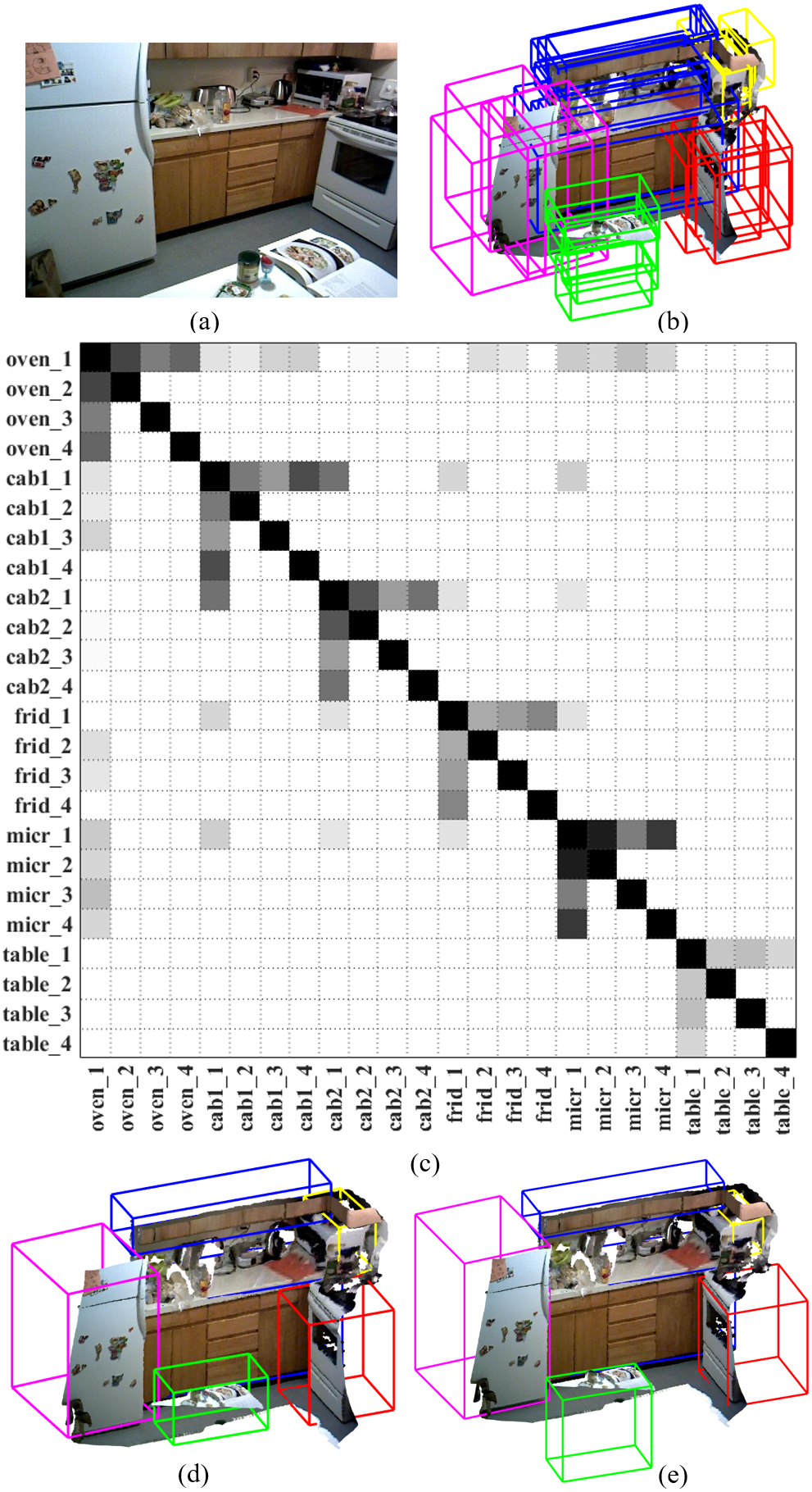}}
	\vspace{-3mm}
	\caption{\label{fig:Relation_vis} Visualization of learned relation graph on one scene of SunRGB-D~\cite{SunRGBD}. (a) RGB image (for better view). (b) 3D bounding box candidates generated by our 3D proposal generation module. (c) One learned relation graph. (d) The final output. (e) Ground truth. The color information of point cloud is only used for better display.}
\end{figure}

\begin{figure*}[t!] 
	\center{\includegraphics[width=1\textwidth]{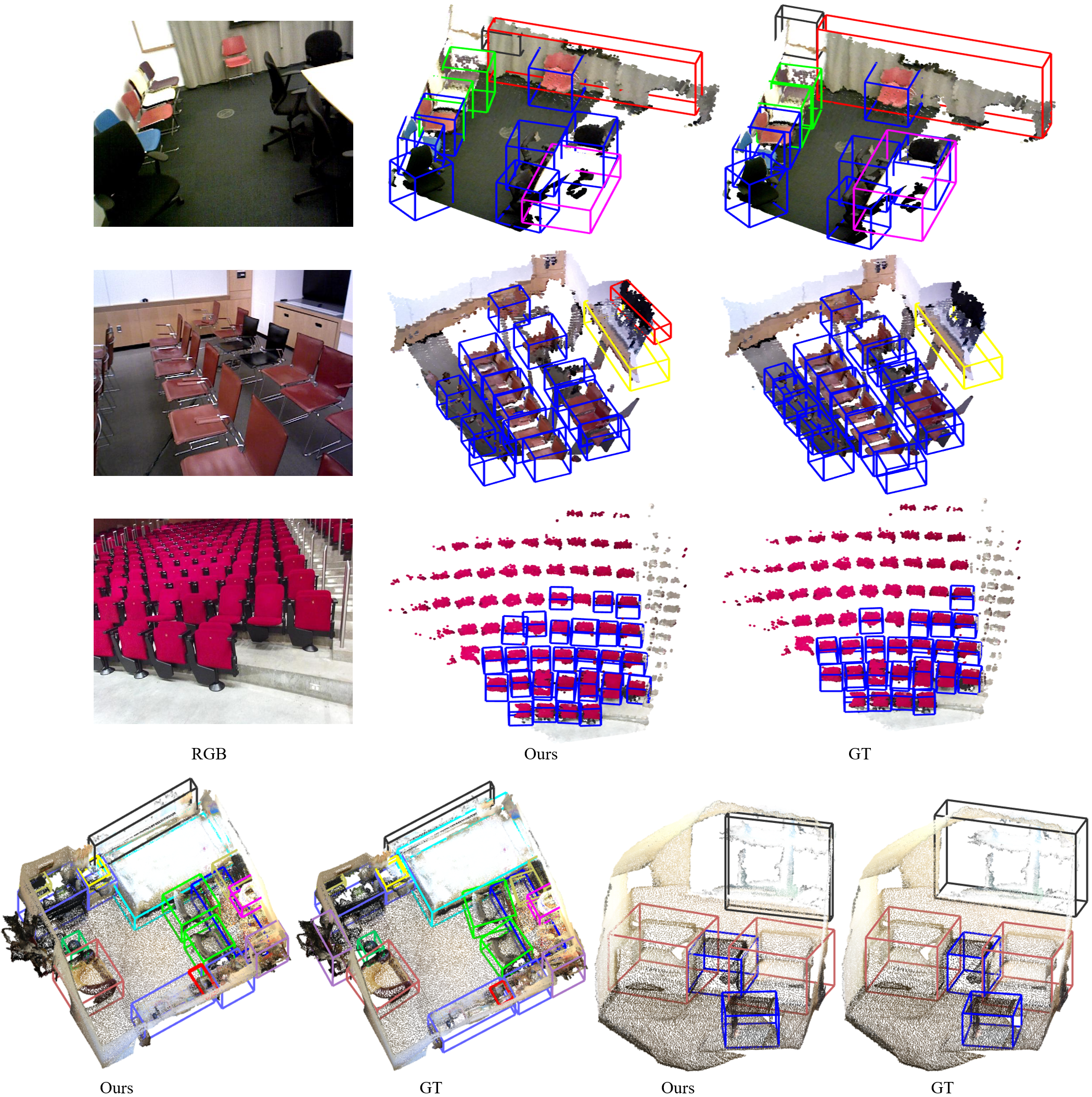}}
	\vspace{-3mm}
	\caption{\label{fig:Detection_vis} Qualitative results of 3D object detection in point cloud, the first three rows are scenes of SunRGB-D~\cite{SunRGBD} and the last row is scenes of ScanNet~\cite{Scannet} dataset. We use the RGB image and color information of point cloud only for better visualization. }
\end{figure*}

\subsection{Comparison with the State of the Art Methods} 
We first evaluate our method on SunRGB-D dataset i.e. ten common 3D object categories. Note that we do not use the color information of point clouds in our model. We report the average precision (AP) with an IoU threshold 0.25 as the evaluation metric. As shown in Table~\ref{SunRGBD_Acc}, our method performs better than the state-of-the-art approaches. The results of baseline methods are taken from the original papers for fair comparison. Particularly, the Deep sliding Shapes (DSS)~\cite{DSS} and COG~\cite{COG} are both voxel based detectors which combine RGB and 3D coordinate information for detection and classification. 2D-driven~\cite{2d_driven} and F-PointNet~\cite{Frustum} rely on the 2D object detectors in the projected images with RGB and 3D coordinate information. VoteNet~\cite{VoteNet} only uses raw point cloud as input with 3D coordinate information. Note that our method outperforms DSS~\cite{DSS}, COG~\cite{COG}, 2D-driven~\cite{2d_driven} and F-PointNet~\cite{Frustum} by at least $5.2\%$ mAP@0.25 even though they use dual modalities. Our method also outperforms VoteNet~\cite{VoteNet}. Furthermore, our method provides the best results on $6/10$ classes, even on objects with partially missing data (e.g., chair and nightstand), and achieves higher accuracy than VoteNet~\cite{VoteNet} on categories such as bathtub, bookshelf, desk and sofa, mainly because our 3D bounding box candidate regression strategy is more stable.

Table~\ref{ScanNet_Acc_detail} shows results on the ScanNet dataset with individual accuracies on the 18 categories. Table~\ref{ScanNet_Acc} shows the overall accuracy. Particularly, 3D-SIS~\cite{3DSIS} uses 3D CNN to detect objects which combines 3D coordinates and multi-views to improve performance. We choose two cases (5 views and 3D coordinates, 3D coordinates only) as the inputs of 3D-SIS for comparison. MRCNN~\cite{3DSIS,MaskRCNN} directly projects the 2D proposals from Mask-RCNN~\cite{MaskRCNN} on 3D point clouds to estimate the 3D bounding boxes. GSPN~\cite{Gspn} uses the Mask-RCNN based framework and PointNet++~\cite{Pointnet++} backbone to generate 3D object proposals which is supervised by the instance segmentation labels. As summarized in Table~\ref{ScanNet_Acc_detail}, our method performs better than 3D-SIS~\cite{3DSIS} and VoteNet~\cite{VoteNet} on $15/18$ classes, and achieves higher accuracy on categories such as chair, table and desk, where the interactions between these objects are complex. For categories with small geometric variations such as door and sink, our method does not achieve the best scores. Our method outperforms all the previous state-of-the-art methods even though it uses only the 3D information.

\subsection{Model Visualization and Qualitative Results}
We show visualizations of a group of 3D bounding box candidates, the relation graph and the final output generated by our model in Figure~\ref{fig:Relation_vis}. We show only one of the three graphs. In Figure ~\ref{fig:Relation_vis} (b) (c), we show only four 3D bounding box candidates per object with higher accuracy. For example, for the candidates of oven, we use oven$\_$1, oven$\_$2, oven$\_$3 and oven$\_$4 to represent them. The Figure~\ref{fig:Relation_vis}(c) is one of the corresponding learned 3 graphs from candidates in Figure~\ref{fig:Relation_vis}(b). In order to highlight objects with strong relationships in Figure~\ref{fig:Relation_vis}(c), we only show the relationship between the candidate numbered 1 and all other candidates numbered 1 for the long distance relationships, and we show the relationship between all four candidates of an object for the local relationships. The darker the color of the square in Figure~\ref{fig:Relation_vis}(c), the stronger the relationship. We can see that there are usually strong interactions between the four candidates of one object, but the long distance relationship between different objects only occasionally exists. For example, there is a weak relationship between the microwave and the cabinet, but the relationships between table and other objects have not been learned. One explanation is that a table is not necessary found in kitchen scenes, hence, its relationship with other objects is difficult to establish.

We show examples of our detection results on SunRGB-D~\cite{SunRGBD} and ScanNet~\cite{Scannet} dataset in Figure~\ref{fig:Detection_vis}. We select 5 challenging scenes which contain partially scanned objects, size changes, occlusions, contact connections, dense placement, and a wide variety of relationships that are difficult to establish. In the first scene of SunRGB-D dataset~\cite{SunRGBD}, our method successfully detect most of the objects, although some objects have a slight deviation from the ground truth, such as curtain and chairs. We can see that our method ignores the whiteboard because it does not have complex geometric information and requires color information to be recognized. The whiteboard also does not have a strong relationship with the surrounding objects. Moreover, the desk is successfully detected because its appearance features can self-reinforce with the information of the surrounding chairs. In the second scene, if only a small part of a chair is scanned, our method can not detect it because too few points do not provide enough information to regress the 3D bounding box. Our method detects the computer that is not in the ground truth. In the third scene, our approach correctly detects more chairs than the annotations given in the ground truth since the partial scanned objects use the information from the surrounding chairs to predict its 3D bounding boxes. 

Detecting thin objects seems to be a limitation of our method. In the scenes of ScanNet~\cite{Scannet}, our method has large errors on very thin objects like window, laptop, and small box, and misses the wardrobe embedded in the wall. Most of these errors occur because we we do not use color information.

\section{Conclusion}
We proposed a relation graph network for 3D object detection in point clouds. Our network jointly learns the pseudo centers and direction vectors from the sampled points on the object surface, which are used to regress 3D bounding box candidates. We introduced a point attention pooling method to adaptively extract uniform and accurate appearance features for each 3D proposal, which benefit from the direction and semantic interactions of interior points. Equipped with the uniform appearance and position features, we built a 3D object-object relation graph to consider the relationships between all 3D proposals. Finally, we exploit the multi graphs and supervised graph strategies to improve the performance of relation graph. Experiments on two challenging benchmark datasets show that our method quantitatively and qualitatively obtains better performance than existing state-of-the-art in 3D object detection.

\section*{Acknowledgment}
This work was supported in part by National Natural Science Foundation of China under Grant 61573134, Grant 61973106 and in part by the Australian Research Council under Grant DP190102443. Thank Yifeng Zhang and Tingting Yang from Hunan University for helping with baseline experiments setup.

\ifCLASSOPTIONcaptionsoff
  \newpage
\fi
\bibliographystyle{IEEEtran}
\bibliography{Reference}

\end{document}